\documentclass{article}

\usepackage[preprint]{neurips_2026}

\usepackage[utf8]{inputenc}
\usepackage[T1]{fontenc}
\usepackage{booktabs}
\usepackage{amsmath,amssymb,amsfonts}
\usepackage{nicefrac}
\usepackage{microtype}
\usepackage{graphicx}
\usepackage{xcolor}
\usepackage{array}
\usepackage{tabularx}
\usepackage{enumitem}
\usepackage{makecell}
\usepackage[bookmarks=false,colorlinks=true,linkcolor=blue,citecolor=blue,urlcolor=blue]{hyperref}
\usepackage{tcolorbox}
\tcbuselibrary{breakable,skins}

\newcommand{\tablenote}[1]{\par\vspace{4pt}\begin{minipage}{0.92\linewidth}\footnotesize\textit{#1}\end{minipage}}

\graphicspath{{images/}}

\title{Evaluating Strategic Reasoning in Forecasting Agents}

\author{
  Tom Liptay \\
  FutureSearch \\
  \texttt{tom@futuresearch.ai}
  \And
  Dan Schwarz \\
  FutureSearch \\
  \texttt{dan@futuresearch.ai}
  \And
  Rafael Poyiadzi \\
  FutureSearch \\
  \texttt{rafael@futuresearch.ai}
  \And
  Jack Wildman \\
  FutureSearch \\
  \texttt{jack@futuresearch.ai}
  \And
  Nikos I. Bosse \\
  FutureSearch \\
  \texttt{nikos@futuresearch.ai}
}

\date{}

\begin{document}

\maketitle

\begin{abstract}
  Forecasting benchmarks produce accuracy leaderboards but little insight into why some forecasters are more accurate than others. We introduce Bench to the Future 2 (BTF-2), 1,417 pastcasting questions with a frozen 15M-document research corpus in which agents reproducibly research and forecast offline, producing full reasoning traces. BTF-2 detects accuracy differences of 0.004 Brier score, and can distinguish differential agent strengths in research vs.\ judgment. We build a forecaster 0.011 Brier more accurate than any single frontier agent, and use it to evaluate agent strategic reasoning without hindsight bias. We find the better forecaster differs primarily in its pre-mortem analysis of its blind spots and consideration of black swans. Expert human forecasters found the dominant strategic reasoning failures of frontier agents are in assessing political and business leaders' incentives, judging their likelihood to follow through on stated plans, and modeling institutional processes.
\end{abstract}

\section{Introduction}
\label{sec:introduction}

There is no clear upper bound on the accuracy of the best forecaster on real-world outcomes. AI forecasting systems have shown continuous improvement since Autocast~\citep{zou2022forecasting} in 2022, but most forecasting benchmarks produce leaderboards without explaining what leads to better accuracy.

Frontier AI forecasters are agents that search and read web pages to research each question. Most efforts to evaluate forecasting agents have used live forecasting questions that require waiting for resolutions to assess accuracy, and hence are not reproducible since new information continually becomes available on the web. Assessments of live forecasts are also subject to hindsight bias when evaluators cannot reconstruct what information was known at the time the forecast was made. Approaches like ForecastBench~\citep{karger2025forecastbench}, the Metaculus AI Benchmarking Series~\citep{metaculusAIBenchmarking2024}, and Prophet Arena~\citep{yang2025prophet} use live forecasting questions.

Approaches that attempt to make forecasting reproducible risk data leakage. \citet{alur2025aia}, for instance, relies on an LLM-judge to detect potential information leakage of the future. Leakage has been documented in the approach used by \citet{phan2024llms}.

The original Bench to the Future (BTF)~\citep{futuresearch2025btfv2} introduced a hermetic offline corpus, a snapshot of tens of thousands of web pages for each forecasting question, captured at question creation time. This ensured that models could not access information after the question is generated, and demonstrated that the forecasts were reproducible over time. However, BTF questions were created in 2025, predating most frontier models' training cutoffs. BTF-2 has 1417 questions that are more challenging, and less correlated, than the 299 questions in BTF.

\section{The BTF-2 Benchmark}
\label{sec:benchmark}

Bench to the Future 2 (BTF-2) is based on a dataset of 1499 questions discussed in \citet{bosse2026automating}, originally asked in October 2025 and resolved in December 2025. Each question has an average of $\approx$10,100 web pages scraped and stored offline (range: 6,000--19,000), for a total of $\approx$16.2 million documents, of which $\approx$8.7 million pages are unique.

We reduced this question set to 1417 questions, removing questions where timing mismatches between generation, scraping, and resolution made them misleading. We applied conservative filters, removing any question with ambiguous resolution criteria, risk of incorrect resolution, or outcomes knowable before the question was posed. The questions span regulatory and policy actions ($\approx$23\%), US government and public policy ($\approx$13\%), macroeconomics and markets ($\approx$12\%), international security and diplomacy ($\approx$11\%), court cases and investigations ($\approx$10\%), and smaller clusters covering Gaza war diplomacy, extreme weather, Iran's nuclear program, commercial space launches, COP30 climate negotiations, and elections.

LLMs with training cutoffs before Oct 2025 can pastcast these questions without contamination risk. (BTF-2 forecasts made by Claude Opus 4.7, which has a January 2026 cutoff, did appear to be using knowledge beyond what was known in Oct 2025.) We use the RetroSearch system described in \cite{futuresearch2025btfv2} to provide Search and Page Read tools to the forecasting agents, approximating live internet search.

\subsection{Validating the statistical power}
The best forecast from \citet{bosse2026automating} used live-web research summaries with Gemini 3.0 Pro Preview and a structured prompt with forecasting tips (the ``Bosse et al.\ prompt''). We re-ran this exact setup---same research summaries, same prompt---with the newer Gemini 3.1 Pro Preview.

\begin{table}[ht]
  \centering
  \caption{Validating detection of model improvements: Gemini 3.1 Pro vs.\ 3.0 Pro.}
  \label{tab:retrosearch-validation}
  \small
  \begin{tabular}{@{}lcccc@{}}
    \toprule
    \textbf{Forecaster} & \textbf{N} & \textbf{Brier} & \textbf{$\Delta$} & \textbf{95\% CI} \\
    \midrule
    Gemini 3.1 Pro (re-run)   & 1417 & \textbf{0.129} & ---   & ---            \\
    Gemini 3.0 Pro (original) & 1417 & 0.138          & 0.009 & [0.002, 0.016] \\
    \bottomrule
  \end{tabular}
  \tablenote{Both forecasters used the same pre-gathered live-web research summaries and the Bosse et al.\ prompt; they differ only in model version. $\Delta$ is the Brier difference relative to the top row. Best Brier is \textbf{bolded}.}
\end{table}

The newer model achieves significantly better accuracy on the same questions with the same approach (Table~\ref{tab:retrosearch-validation}). This is consistent with Gemini 3.1 Pro being an overall more intelligent model.

\subsection{Validating the question difficulty}
To validate that BTF-2 questions are genuinely difficult, but tractable, we measured divergence across repeated independent rollouts of a strong forecasting agent.

We selected 200 questions that live (prospective) Gemini 3 Pro runs from \citet{bosse2026automating} had identified as difficult, based on high Brier scores and inter-run variance. Table~\ref{tab:question-difficulty} shows how many of 8 independent rollouts were on the correct side of the 50\% threshold, when run by an Opus 4.6 agent, the strongest single forecasting agent as shown later in Table~\ref{tab:single-prompt}.

\begin{table}[ht]
  \centering
  \caption{Opus 4.6 Agent rollout consistency on 200 difficult questions.}
  \label{tab:question-difficulty}
  \begin{tabular}{lccccc}
    \toprule
    Directionally correct rollouts (of 8) & 0 & 1--3 & 4 & 5--7 & 8 \\
    \midrule
    Questions & 89 (45\%) & 36 (18\%) & 5 (3\%) & 34 (17\%) & 36 (18\%) \\
    \bottomrule
  \end{tabular}
  \tablenote{Directionally correct means a forecast of <50\% if the resolution was NO, and >50\% if the resolution was YES.}
\end{table}

On 38\% of questions, for some but not all of the 8 rollouts, Opus produced forecasts on both sides of the 50\% threshold (mean per-question $\sigma = 0.08$), suggesting the forecasting tasks were difficult but tractable. Of the remainder, there are at least three possibilities that we cannot distinguish: (1) The critical information was not in the RetroSearch corpus, (2) The forecast was too difficult for the agent, or (3) the result was genuinely unlikely e.g. a forecast of 20\% would be excellent given the information available, but the event in fact occurs, as it should one in five times.

\section{Methodology}
\label{sec:methods}

Forecasting attempts in BTF-2 produce a full sequence of searches, page reads, thoughts, and a final rationale, and can be re-run many times. This enables a series of experiments on the research process and strategic reasoning of the forecasters, in addition to experiments on accuracy.

\paragraph{Agent evaluation.}
We aim to determine: (1)~Which frontier LLM makes the most accurate forecasting agent, (2)~How much the best agent's advantage come from research strategy or from judgment, (3)~How frontier agents compare to a state-of-the-art forecaster.

In all experiments with forecasting agents, we evaluate agents run by frontier models at high effort with training window cutoffs before Oct 2025, from their commercial APIs. All agents use a ReAct-style architecture~\citep{yao2023react}. Specifically, we use FutureSearch's ReAct implementation~\citep{futuresearch2026agents}, whose system prompt, toolkit, and time-management mechanism were tuned against Deep Research Bench~\citep{bosse2025deepresearchbench}, a benchmark of challenging web-research tasks. Accordingly we gave the ReAct agents an iteration budget of 10, so up to 10 times, agents can call a combination of parallel Search and Read tools. A typical agent makes $\approx$10--20 tool calls, takes about 5--6 minutes total, processes $\approx$80--200k input and output tokens, looks at 50--300 search snippets, reads 5--20 pages in full, and vary in cost from \$0.14 (Grok 4.2 agent) to \$0.55 (Opus 4.6 agent). For Q(2), we compare agents to models given pre-gathered research.

\paragraph{Reasoning patterns across different forecasting agents.}
To understand what distinguishes better forecasting agents, we score each forecast rationale against Tetlock's CHAMPS KNOW framework~\citep{tetlock2015superforecasting, chang2016developing}, a taxonomy 10 dimensions associated with better human forecasting accuracy. For each rationale, a Gemini 3.1 Pro agent ranks the 10 dimensions from 1 (most prominent) to 10 (least prominent). We compare dimension profiles across models and between individual forecasting agents and the SOTA forecaster (Section~\ref{sec:taxonomy}).

\paragraph{Strategic reasoning mistakes of the best forecasting agents.}
To understand the biggest weaknesses of the best forecasting agents, we had expert human forecasters examine a sample of forecasts to judge whether the failures were due to a significant error in judgment---one that an expert human forecaster would be unlikely to make.

To isolate LLM judgment from scaffold effects, we used the best single agent from Section~\ref{sec:single-prompt} (no ensembling, no forecasting guidance).

To ensure the failures were not simply human hindsight bias from knowing the outcome, in addition to looking at the worst $\approx$5\% by absolute accuracy score, we also looked at the worst $\approx$5\% as measured by the accuracy difference between that no-guidance forecasting agent and the SOTA forecasting agent from Section~\ref{sec:best-forecaster}, for a total of 130 questions.

To verify reproducibility, we re-ran the agent and retained only cases where both runs made the same error.

\section{Results}
\label{sec:results}

\subsection{Comparing Frontier Agents}
\label{sec:single-prompt}

We first evaluate agents with no guidance on forecasting practices, approximating the experience of asking ChatGPT, Claude, or Gemini on a standard ``pro'' subscription tier.

We evaluated Anthropic's Claude Opus 4.6 (thinking: ``adaptive'', effort: ``high''), Google's Gemini 3.1 Pro Preview (thinking\_level: ``high''), OpenAI GPT-5.4 (reasoning\_effort: ``high''), and xAI's Grok 4.20 Beta. Every agent was given the following straightforward prompt:

\begin{quote}
\emph{``You have been given a prediction question with its resolution criteria. Your task is to research this question and produce the most accurate probabilistic forecast you can. Write a brief rationale summarizing your research and reasoning, then provide your final forecast as a probability between 0 and 100.''}
\end{quote}

Table~\ref{tab:single-prompt} shows the resulting Brier scores. $\Delta$ corresponds to the difference in Brier score with Opus 4.6 on the same set of questions. 95\% CI is calculated with bootstrapping throughout the paper. Calibration and refinement are calculated as in \citet{bosse2026automating}.

\begin{table}[tbp]
  \centering
  \caption{Frontier agent accuracy on BTF-2.}
  \label{tab:single-prompt}
  \small
  \setlength{\tabcolsep}{5pt}
  \begin{tabular}{@{}lcccccc@{}}
    \toprule
    \textbf{Agent} & \textbf{N} & \textbf{Brier} & \textbf{$\Delta$} & \textbf{95\% CI} & \textbf{Calibration} & \textbf{Refinement} \\
    \midrule
    Opus 4.6 Agent           & 1417 & \textbf{0.131} & ---   & ---              & 0.005 & \textbf{0.073} \\
    Gemini 3.1 Pro Agent     & 1417 & 0.143          & 0.012 & [0.002, 0.021]   & 0.012 & 0.067 \\
    GPT-5.4 Agent            & 1417 & 0.151          & 0.020 & [0.012, 0.029]   & 0.010 & 0.057 \\
    Grok 4.20 Beta Agent     & 1300 & 0.165          & 0.033 & [0.022, 0.044]   & \textbf{0.003} & 0.039 \\
    \bottomrule
  \end{tabular}
  \tablenote{All agents use the simple prompt (Section~\ref{sec:single-prompt}) with no forecasting guidance. $\Delta$ is the Brier difference relative to Opus 4.6 Agent. Best value in each column is \textbf{bolded}. Grok ran on $N = 1300$ due to technical failures on 117 questions.}
\end{table}

The Opus 4.6 agent's superior accuracy over the other 3 leading agents is statistically significant at the 95\% confidence level. The other three agents have a clear ordering in accuracy, with Gemini 3.1 Pro second, followed by GPT-5.4 and then Grok 4.20 Beta.

Brier scores decompose into calibration and refinement. Calibration measures alignment between forecast probabilities and actual frequencies (lower is better), reflecting how well a forecaster knows the limits of its knowledge. Refinement measures ability to differentiate between outcomes (higher is better), reflecting how much useful information a forecaster brings to bear.

Opus 4.6 is better calibrated than Gemini 3.1 Pro and GPT-5.4. Surprisingly, we find that the Grok 4.20 agent is the best calibrated among the agents, despite having the worst accuracy. This is the equivalent of a meteorologist who always forecasts the base rate for rain in London, never looking at a radar. They are perfectly calibrated, but have zero refinement.

Refinement rankings match the accuracy ordering.

\subsection{Isolating Judgment from Research}
\label{sec:research-informed}

We investigated whether the Opus 4.6 agent's lead comes from a superior research strategy in searching and page reading, or in superior judgment over that evidence. We gave all four models the pre-gathered research summaries from \citet{bosse2026automating}, with no ability to search further, and this simple prompt:

\begin{quote}
\emph{``Based on the information provided, forecast the probability of a `YES' outcome as a percentage between 0 and 100. Write a brief rationale, then provide your final answer.''}
\end{quote}

\begin{table}[tbp]
  \centering
  \caption{Judgment over fixed evidence.}
  \label{tab:research-informed}
  \small
  \setlength{\tabcolsep}{5pt}
  \begin{tabular}{@{}lcccccc@{}}
    \toprule
    \textbf{Model} & \textbf{N} & \textbf{Brier} & \textbf{$\Delta$} & \textbf{95\% CI} & \textbf{Calibration} & \textbf{Refinement} \\
    \midrule
    Gemini 3.1 Pro & 1417 & \textbf{0.141} & 0.012 & [0.006, 0.018]   & \textbf{0.012} & \textbf{0.069} \\
    Opus 4.6           & 1417 & 0.153          & 0.024 & [0.015, 0.033]   & 0.015 & 0.061 \\
    GPT-5.4            & 1417 & 0.156          & 0.027 & [0.019, 0.034]   & 0.026 & 0.068 \\
    Grok 4.20 Beta     & 1417 & 0.163          & 0.033 & [0.025, 0.042]   & 0.020 & 0.056 \\
    \bottomrule
  \end{tabular}
  \tablenote{Models were given pre-gathered research summaries with no ability to search further. $\Delta$ is the Brier difference relative to the best single-model forecast (0.129; Table~\ref{tab:retrosearch-validation}). Best value in each column is \textbf{bolded}.}
\end{table}

Forecasting agents with this pre-compiled research did significantly worse than when conducting their own research. Opus 4.6 declined from 0.131 to 0.153, with worse calibration and refinement, indicating its agent research finds information the pre-gathered summaries miss.

Refinement improved for Gemini 3.1 Pro, GPT-5.4, and Grok 4.20, suggesting the Bosse et al.\ research exceeded what those agents found independently. Calibration declined for all models, for unclear reasons.

We also tested Opus 4.6 as an agentic researcher with the elaborate Bosse et al.\ prompt. Table~\ref{tab:conditions-summary} summarizes the best Brier scores under each combination of agent does research vs. research provided, and simple prompt vs. prompt with forecasting best practices.

\begin{table}[tbp]
  \centering
  \caption{Best Brier score by research source and prompt.}
  \label{tab:conditions-summary}
  \small
  \begin{tabular}{@{}lcc@{}}
    \toprule
    & \textbf{Simple Prompt} & \textbf{Bosse et al.\ Prompt} \\
    \midrule
    \textbf{Agent does research}  & Opus 4.6 (0.131)   & Opus 4.6 (0.131)   \\
    \textbf{Bosse et al.\ research provided} & Gemini 3.1 Pro (0.141) & Gemini 3.1 Pro (0.129) \\
    \bottomrule
  \end{tabular}
\end{table}

The Gemini 3.1 Pro re-run of the Bosse et al.\ method (Table~\ref{tab:retrosearch-validation}) remains the most accurate single-model forecast. But BTF-2 evaluates agents conducting their research, the first row above, under which Opus 4.6 is best under both simple and elaborate agent prompts. Two examples of the full trace of the Opus 4.6 agent researching and forecasting are in the Appendix~\ref{app:asuu}.

\subsection{State-of-the-art forecasting on BTF-2}
\label{sec:refinement}
\label{sec:best-forecaster}

BTF-2's reproducibility enabled us to build a state-of-the-art (SOTA) forecasting agent substantially more accurate than any frontier agent. Its rationales provide a hindsight-bias-free standard for analyzing where frontier agents differ (Section~\ref{sec:taxonomy}) and fail (Section~\ref{sec:examples}).

Accuracy-improving approaches include: (1) Taking the mean of multiple runs of the same agent (``wisdom of the crowd''); (2) Providing the agents with forecasts to related questions; (3) Re-calibrating forecasts based on historic calibration curves, (4) Asking a longer-scope version of the question first, and (5) Detecting and correcting for biases in forecast rationales.

\begin{table}[tbp]
  \centering
  \caption{SOTA and component forecasting agent performance.}
  \label{tab:refinement}
  \small
  \setlength{\tabcolsep}{5pt}
  \begin{tabular}{@{}lccccccc@{}}
    \toprule
    \textbf{Forecasting Agent} & \textbf{N} & \textbf{Brier} & \textbf{$\Delta$} & \textbf{95\% CI} & \textbf{Calibration} & \textbf{Refinement} & \textbf{Cost} \\
    \midrule
    SOTA Forecasting Agent    & 1367 & \textbf{0.119} & ---   & ---                & \textbf{0.002} & \textbf{0.081} & \$1.68 \\
    \midrule
    Opus 4.6 Agent           & 1367 & 0.130          & 0.011 & [0.006, 0.017]     & 0.005 & 0.075 & \$0.55 \\
    Opus 4.6 Agent run 2     & 1367 & 0.130          & 0.011 & [0.006, 0.017]     & 0.005 & 0.074 & \$0.55 \\
    Gemini 3.1 Pro Agent     & 1367 & 0.141          & 0.023 & [0.015, 0.031]     & 0.012 & 0.069 & \$0.30 \\
    GPT-5.4 Agent            & 1367 & 0.152          & 0.034 & [0.025, 0.042]     & 0.010 & 0.056 & \$0.24 \\
    \midrule
    \emph{Mean 4 agents}   & \emph{1367} & \emph{0.125} & \emph{0.007} & \emph{[0.002, 0.011]} & \emph{0.007} & \emph{0.081} & --- \\
    \bottomrule
  \end{tabular}
  \tablenote{$\Delta$ is the Brier difference relative to the SOTA Forecasting Agent. Best value in each column is \textbf{bolded}. $N = 1367$ (50 questions excluded due to technical issues in multi-agent runs). Cost is the average agent cost per question, including LLM API calls, searches, snippet generation, and page reads.}
\end{table}

The mean of four agents outperforms any individual (Table~\ref{tab:refinement}), showing a wisdom-of-crowds effect. The SOTA forecasting agent did significantly better than this mean.

The SOTA forecasting agent is also better calibrated than the frontier agents. Frontier agents have a more pronounced sag to them, similar to what is found on Metaculus or prediction markets with human forecasters.

\begin{figure}[tbp]
  \centering
  \includegraphics[width=0.66\textwidth]{calibration_curves.png}
  \caption{Calibration curves for the top three frontier agents and the SOTA forecaster ($N = 1367$). Dashed diagonal indicates perfect calibration. The SOTA forecaster exhibits markedly better calibration, closely tracking the diagonal. Brier scores in parentheses.}
  \label{fig:calibration-curves}
\end{figure}

\subsection{What better forecasting agents focus on}
\label{sec:taxonomy}

The SOTA forecasting agent from Section~\ref{sec:best-forecaster} makes this comparison possible. Because it is significantly more accurate than any single frontier agent, differences in what it emphasizes reflect what better forecasters focus on, without the risk of hindsight bias that would arise from knowing the outcomes.

To characterize these differences, we use Tetlock's CHAMPS KNOW framework~\citep{tetlock2015superforecasting}, a set of 10 dimensions developed in the Good Judgment Project. Training human forecasters on these principles improved accuracy by 6--12\% in a randomized trial~\citep{chang2016developing}. We used Gemini 3.1 Pro to rank-order dimensions by prominence in each rationale. Table~\ref{tab:champs-know} shows top-3 frequency.

\begin{table}[tbp]
  \centering
  \caption{CHAMPS KNOW top-3 frequency by forecasting agent.}
  \label{tab:champs-know}
  \small
  \setlength{\tabcolsep}{4pt}
  \begin{tabular}{@{}clrrrr@{}}
    \toprule
    \textbf{Dim} & \textbf{Description} & \textbf{\makecell{SOTA\\Agent}} & \textbf{\makecell{Opus 4.6\\Agent}} & \textbf{\makecell{Gemini 3.1\\Pro Agent}} & \textbf{\makecell{GPT-5.4\\Agent}} \\
    \midrule
    N & Norms \& Protocols    & 63.1\% & 50.7\% & 57.6\% & 55.7\% \\
    K & Know power players    & 43.2\% & 33.6\% & 37.4\% & 16.4\% \\
    C & Comparison Classes    & 38.9\% & 34.1\% & 36.5\% & 24.4\% \\
    H & Hunt for info         & 38.0\% & 94.3\% & 84.4\% & 97.4\% \\
    P & Pre/Post-mortem       & 37.8\% &  9.5\% &  4.3\% &  6.8\% \\
    S & Select right Qs       & 28.1\% & 39.6\% & 40.3\% & 67.5\% \\
    M & Math/stats models     & 22.6\% & 26.5\% & 25.0\% & 18.9\% \\
    O & Other perspectives    & 20.3\% &  5.1\% &  1.7\% &  1.6\% \\
    A & Adjust \& update      &  5.1\% &  5.9\% & 12.1\% & 10.8\% \\
    W & Wildcards             &  2.9\% &  0.7\% &  0.7\% &  0.3\% \\
    \bottomrule
  \end{tabular}
  \tablenote{Values show the percentage of rationales in which each dimension ranked in the top 3. Agent labels correspond to the agents from Table~\ref{tab:single-prompt}.}
\end{table}

\begin{figure}[tbp]
  \centering
  \includegraphics[width=0.90\textwidth]{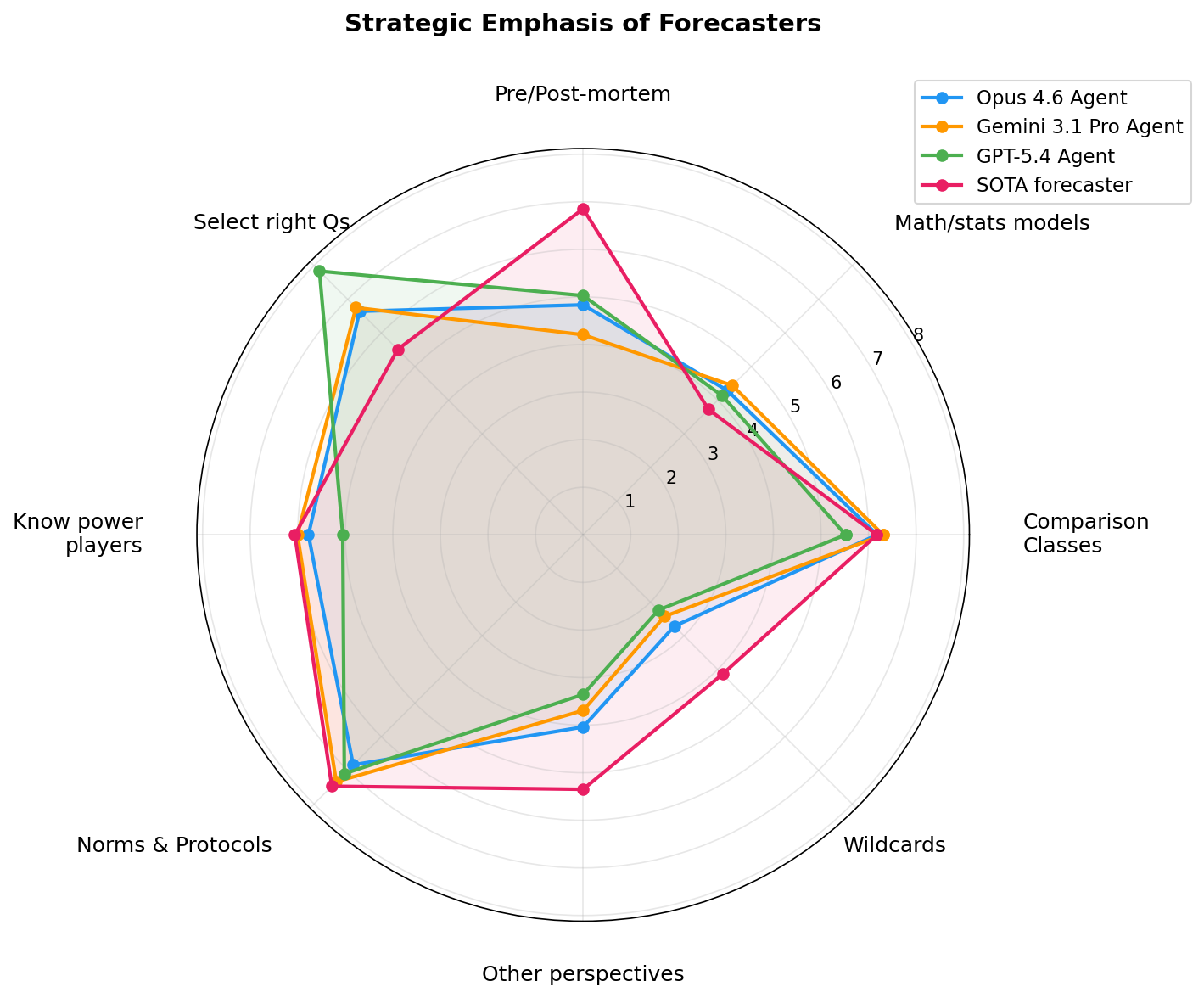}
  \caption{CHAMPS KNOW radar plot showing mean Borda score per dimension (rank 1 = 10 points, rank 10 = 1 point), excluding Hunt for Info and Adjust \& Update which are uninformative for comparison. The SOTA forecasting agent emphasizes Pre/Post-mortem analysis, Other perspectives, and Wildcards far more than any individual frontier model agent.}
  \label{fig:champs-radar}
\end{figure}

Rationales cannot include all forecasting considerations. The SOTA forecasting agent also has a designed structure in its rationale that is not present in the other agents, each of which also vary in structure. Hence this ranking of factors reflects strategy, prioritization, and also conventions of emphasis, and is hence subjective. In particular, human expert forecasters concluded this led to unrealistic comparisons in two of the factors: Hunt for Info, and Adjust and Update, so removed them from the direct comparison radar chart in Figure~\ref{fig:champs-radar}.

The biggest differences in the remaining 8 factors between the SOTA forecasting agent and the less accurate forecasting agents were Pre/Post-mortem, Other perspectives, and Wildcards. Pre/Post-mortem analysis refers to whether the forecasting agent considers, if its forecast ends up being wrong, the most likely reasons why this would occur. Other perspectives and Wildcards, taken together, refer to black swans, unknown unknowns, and blind spots.

All three are epistemic. The primary gap is awareness of uncertainty and the limits of one's knowledge.

\subsection{Strategic reasoning mistakes of the best forecasting agents}
\label{sec:examples}

Expert human forecasters examined runs from the Opus 4.6 forecasting agent with the simple prompt from Section~\ref{sec:single-prompt}, reflecting the best accuracy without any special guidance on forecasting best practices. We selected the least accurate $\approx$5\% of questions overall, and the least accurate $\approx$5\% compared to the SOTA forecasting agent from Section~\ref{sec:best-forecaster}, for a total of 130 questions. This ensures examined failures reflect achievable improvements, not merely surprising outcomes.

Even on these 130 questions, experts judged most forecasts high quality. On 24 of these 130 questions, they judged there were significant strategic reasoning errors. These were broadly in two categories: reasoning errors specific to the forecasting task, and reasoning errors in strategic foresight and world modeling.

Errors specific to the forecasting tasks can be broadly classified as one of these three categories: (1) Considering the most extreme version of an outcome, showing why that extreme outcome was unlikely, and downweighting the likelihood of a milder outcome; (2) Extrapolating from base rates too strictly even when the process leading to that base rate has been disrupted; (3) Producing a great forecast but then falling back to a more conservative prior.

Errors in strategic foresight and world modeling generally fell into two categories: (1) Failing to judge in which cases political or business leaders will follow through on their stated position; and (2) failing to understand the primary incentives faced by political or business leaders.

We include one case study of each of these two types of strategic errors. Full details are provided in Appendix~\ref{app:asuu} and Appendix~\ref{app:cop30}.

\subsubsection{Case Study: Judging when a political leader will follow through on their stated position}
\label{sec:case-asuu}

\textbf{Question:} ``Will ASUU (Academic Staff Union of Universities, Nigeria) declare a nationwide university strike lasting at least 7 consecutive days?'' (Oct 15--Dec 31, 2025)

\medskip
\noindent\textit{Opus 4.6 Forecasting Agent: 72\% / 75\% \quad SOTA Forecasting Agent: 30\% \quad Resolution: No}

\medskip
ASUU had begun a two-week warning strike on October 13, two days before the window opened, as part of a publicly stated escalation sequence: 14-day ultimatum, then two-week warning strike, then indefinite strike. The Opus 4.6 agent gave 72\%, anchoring on this sequence and on a quote from ASUU National President Chris Piwuna that the next phase ``will be total and there will be no going back,'' which it positioned as its primary evidence. In fact, ASUU's National Executive Council suspended the warning strike on October 22, giving the government a one-month window. When rumors of a November 21 indefinite strike circulated, ASUU's national office explicitly denied them. On December 23, ASUU and the Federal Government signed a landmark renegotiated agreement, the first since 2009, including a 40\% salary uplift. No qualifying strike occurred.

The Opus 4.6 agent's error was treating Piwuna's maximalist rhetoric as a commitment rather than as bargaining leverage. The same press conference contained an explicit hedge, ``we will meet after the expiration to decide when to begin an indefinite and comprehensive strike action,'' reserving the very decision the forecasting agent treated as already made. The SOTA forecasting agent at 30\% prospectively identified structural signals pointing toward de-escalation: active negotiations had produced progress on five of seven demands; ASUU typically affords the government a grace period of several weeks after suspending a warning strike; the Nigeria Labour Congress's parallel four-week ultimatum provided a face-saving clock; and Nigerian universities traditionally close for Christmas. The Opus 4.6 agent cited the negotiation progress but discounted it under the escalation-rhetoric frame, and never considered the grace-period pattern or academic-calendar seasonality.

Agent trace analysis confirmed that the Opus forecasting agent encountered and paraphrased Piwuna's hedge (``the union will meet again after the two-week period''), but in its final analysis, it dropped this conditionality.

\subsubsection{Case Study: Modeling an actor's incentives}
\label{sec:case-cop30}

\textbf{Question:} ``Will the C\^{a}mara dos Deputados (Brazilian House of Representatives) Plenary approve the lead proposition containing PL\,1.874/2022 (National Circular Economy Policy) between 2025-10-15 and 2025-12-31?''

\medskip
\noindent\textit{Opus 4.6 Forecasting Agent: 30\% / 35\% \quad SOTA Forecasting Agent: 70\% \quad Resolution: Yes}

\medskip
Brazil's circular economy bill (PL\,1.874/2022) had been repeatedly scheduled for plenary votes throughout 2025 but never voted on. The forecasting agent gave 30\%, anchoring on this pattern of scheduling without action, pending procedural motions, and industry opposition calling the substitute text a ``legislative monster.'' It cited Finance Minister Haddad's statement that the bill was a government priority, but treated that as one factor among many. On October 29, the C\^{a}mara approved the bill, 12 days before COP30 opened in Bel\'{e}m. Brazil's Frente Parlamentar Ambientalista had explicitly delivered a document to the Chamber presidency titled ``The Urgent Legacy That the Legislative Branch Needs to Leave for Brazil at COP30,'' listing the circular economy bill as a priority.

The Opus 4.6 agent never asked why the Lula government would want this chronically stalled bill passed now. The answer was that Brazil was hosting COP30, the UN climate summit, in November. This transformed the bill from a slow procedural grind into a must-ship showcase item for the host country. The word ``COP30'' does not appear anywhere in the Opus 4.6 agent's rationale. The SOTA forecasting agent at 70\% prospectively identified COP30 as ``an exceptionally strong political catalyst''. The Opus 4.6 agent weighted the September 11 industry complaints heavily despite its own evidence of a September 24 breakthrough agreement meant to resolve those very complaints.

Agent trace analysis of the first Opus 4.6 agent run showed 17 web searches focused on bill numbers and procedural status, but no query contained ``COP30,'' ``climate,'' or related. The second Opus 4.6 agent run happened to find COP30 information through a broader query, yet the final synthesis still gave only 35\%, heavily discounting these signals in favor of the procedural base rate.

\section{Discussion}
\label{sec:discussion}

BTF-2 is the first forecasting benchmark with a large number of difficult, diverse questions, and a hermetic corpus for reproducibility. It can distinguish small accuracy differences, capturing differential strengths between strong frontier agents at statistical significance. The performance of frontier agents on BTF-2 also validate the pastcasting methodology and RetroSearch agent toolkit, achieving higher accuracy than agents that had access to the live web. This reproducibility, along with the detailed agent traces, enable experimentation to develop stronger forecasters, without the hindsight bias that comes from using the question resolutions as feedback.

The stronger forecaster used in this paper has a 0.011 Brier score improvement over the Opus 4.6 agent, the most accurate frontier agent not given forecast-specific guidance. To put this in context, a 0.010 Brier improvement is equivalent to a 50 percentage point advantage on 1 out of every 25 questions, or a 10 percentage point advantage on every question. Though expert human forecasters judge that the Opus 4.6 agent is a great forecaster, this Brier difference points to significant headroom.

Investigations into agent traces and rationales offer insights into this headroom. The CHAMPS KNOW analysis finds that better forecasters are distinguished by epistemics, particularly pre-mortem analysis, correcting for blind spots, and considering black swans. The case studies show that forecasting agents sometimes fail to model the incentives of political and business leaders, treating stated positions as commitments rather than strategic moves. This suggests paths forward to build better agentic forecasters, as well as the methodology to evaluate that progress.

\paragraph{Limitations}
BTF-2 covers one time period (Oct--Dec 2025), so cannot evaluate models with a more recent training window cutoff. The RetroSearch methodology requires periodically producing new benchmarks.

BTF-2 is weighted toward geopolitics, policy, and macroeconomics. The finding that the most serious failures are in incentive-modeling may partly reflect question composition.

The better accuracy and calibration of the SOTA forecaster may reflect overfitting to the BTF-2 questions.

The CHAMPS KNOW analysis is done by a single LLM call on final rationales, not full traces. As the expert human forecast reviews uncovered, evaluating the full series of searches, page reads, and thoughts of each agent can add information that rationales alone lack.

Our SOTA forecaster wrote a structured rationale that likely influences the CHAMPS KNOW ranks. In contrast, the single forecasting Agents did not have any imposed structure on their rationales.

While BTF-2 does enable controlled experiments in reasoning strategies for agents, causally isolating the role of strategic factors is challenging. Prompting agents to pay more or less attention to individual factors does not ensure they actually do this.

We applied expert human review on only 130 questions of 1417, due to costs and time. This covered all the questions with low accuracy scores for Opus 4.6 agents, as well as divergence between the Opus 4.6 agent and the SOTA forecaster. But it neglects questions where the agent was "right for the wrong reason". A more complete analysis would likely find strategic reasoning failures in questions with high accuracy scores.

\paragraph{Future Work}

A natural followup to the next version of BTF is to include conditional forecasting questions. These may be more useful in strategic decision making, and also are significantly harder, allowing more exploration of the room between the best agents.

\begin{ack}
The authors have no funding or competing interests to disclose.
\end{ack}

\bibliographystyle{plainnat}
\bibliography{references}

\clearpage
\appendix

\section{Full Case Study Details}
\label{app:case-details}

This appendix provides the complete forecasting question, resolution criteria, and a verbatim reconstruction of the Opus 4.6 agent's research process for each case study discussed in Section~\ref{sec:examples}. The traces are reconstructed from observation logs and show every web search query, page read, and intermediate reasoning turn exactly as produced by the agent. The agent's final rationale appears inline as the last turn of each trace.

\subsection{ASUU Strike}
\label{app:asuu}

\subsubsection*{Forecasting Question}

\begin{tcolorbox}[colback=gray!5, colframe=gray!50]
\small
Between 00:00 WAT Oct 15 and 23:59 WAT Dec 31, 2025, will ASUU declare a nationwide university strike lasting at least 7 consecutive days?
\end{tcolorbox}

\subsubsection*{Resolution Criteria}

\begin{tcolorbox}[colback=gray!5, colframe=gray!50, breakable]
\footnotesize
\textbf{Forecasting target:} Resolve YES if there is any single episode, initiated during the window from 00:00 WAT on 2025-10-15 through 23:59 WAT on 2025-12-31 (inclusive), in which ASUU's national leadership declares a nationwide strike that remains in effect for at least 168 uninterrupted hours (7 consecutive days). Resolve NO otherwise.

\medskip
\textbf{Definitions and scope:}
\begin{itemize}[nosep,leftmargin=1.5em]
\item \textit{ASUU:} Academic Staff Union of Universities, Nigeria. ``ASUU's national leadership'' means the National Executive Council (NEC) or national officers acting on behalf of the union (e.g., the National President), consistent with reporting on the September 2025 ultimatum.
\item \textit{Strike/industrial action:} A formally declared work stoppage by academic staff represented by ASUU. A ``strike'' includes a ``warning strike'' or an ``indefinite strike'' if declared by ASUU's national leadership.
\item \textit{Nationwide:} The declaration is not limited to a specific institution, zone, or subset; it is announced to apply across universities nationally (e.g., described as ``nationwide,'' ``national,'' ``across universities nationwide''). It must at least cover ASUU chapters in Nigeria's federal universities. It is not necessary to verify compliance by every campus; the declaration's stated scope controls.
\item \textit{Consecutive days:} Counted as 168 continuous hours from the declared/reported strike start time until the strike's suspension or expiration is officially announced. Weekends and public holidays count toward the total. If the strike is suspended or called off before 168 hours elapse, the episode does not qualify.
\item \textit{Time window:} Only strike episodes that are initiated (i.e., declared to begin) between 00:00 WAT on 2025-10-15 and 23:59 WAT on 2025-12-31 are eligible. A strike that starts before 2025-10-15 but continues into the window does not qualify; a strike that starts within the window and continues beyond 2025-12-31 can still qualify so long as it reaches 168 hours within the window.
\item \textit{Exclusions:} Actions limited to a single university/zone; actions solely by other unions (e.g., SSANU/NASU) without an ASUU national strike declaration; isolated boycotts, protests, or sit-ins absent a strike declaration by ASUU national leadership; or ``no work'' days without an overarching national strike declaration.
\end{itemize}

\medskip
\textbf{Evidence required to resolve:}
\begin{itemize}[nosep,leftmargin=1.5em]
\item \textit{Primary evidence:} An official ASUU national announcement (e.g., press release on its official channels such as website or verified social media) declaring a nationwide strike and stating a start time/date.
\item \textit{Corroboration:} Reports by at least two reputable Nigeria-based outlets confirming the nationwide strike and its start date (e.g., Premium Times, Punch, Vanguard, The Nation, The Guardian Nigeria, Channels TV).
\end{itemize}

\medskip
\textbf{Adjudication notes:}
\begin{itemize}[nosep,leftmargin=1.5em]
\item \textit{Start time:} If ASUU's announcement provides a start time/date, use that; otherwise, use the earliest start time/date reported by at least two independent outlets.
\item \textit{End time/duration:} If ASUU announces a suspension/call-off, use that time; otherwise, if at least two independent outlets report the strike remained in effect without suspension for 168 hours from the start, that satisfies duration. Partial exemptions (e.g., essential services) do not break continuity if the strike remains officially in effect.
\item \textit{Multiple episodes:} If multiple strike episodes occur, a single qualifying episode suffices for YES.
\item \textit{Time zone:} All times are WAT (UTC+01:00).
\end{itemize}
\end{tcolorbox}

\subsubsection*{Agent Research Trace}

\begin{tcolorbox}[colback=white, colframe=gray!40, breakable]
\footnotesize

\textbf{Turn 1} \\[2pt]
\textbf{\textsc{Thought}:}
\begin{itemize}[nosep,leftmargin=1.5em]
\item The forecast must evaluate whether the strike is initiated (begins) specifically between 00:00 WAT on 2025-10-15 and 23:59 WAT on 2025-12-31.
\item The strike must be declared by ASUU's national leadership (NEC or national officers), not local chapters or other unions.
\item The strike must be nationwide in scope, specifically including Nigeria's federal universities.
\item The strike must last for at least 168 consecutive hours (7 full days) to meet the resolution criteria.
\item The forecast must account for the exclusion of any strike that starts before 2025-10-15, even if it continues into the window.
\item The final forecast must be provided as a probability between 0 and 100.
\item The response must include a rationale summarizing research and reasoning for the forecast.
\end{itemize}

\tcbline

\textbf{Turn 2} \\[2pt]
{\color{blue!70}\textbf{\textsc{Search}}:} \texttt{ASUU nationwide strike October 2025 Nigeria} \\
{\color{blue!70}\textbf{\textsc{Search}}:} \texttt{ASUU ultimatum September 2025 warning strike} \\
{\color{blue!70}\textbf{\textsc{Search}}:} \texttt{ASUU strike 2025 latest news Nigeria}

\tcbline

\textbf{Turn 3} \\[2pt]
{\color{blue!70}\textbf{\textsc{Search}}:} \texttt{ASUU warning strike October 13 2025 end date indefinite strike} \\
{\color{blue!70}\textbf{\textsc{Search}}:} \texttt{ASUU indefinite strike after warning strike October 2025} \\[2pt]
{\color{green!50!black}\textbf{\textsc{Read}}:} channelstv.com --- \textit{ASUU Declares Two-Week Warning Strike} \\
{\color{green!50!black}\textbf{\textsc{Read}}:} guardian.ng --- \textit{ASUU confirms talks with FG's team as strike enters seventh day} \\
{\color{green!50!black}\textbf{\textsc{Read}}:} vanguardngr.com --- \textit{ASUU Strike: Union reports progress in talks with FG}

\tcbline

\textbf{Turn 4} \\[2pt]
\textbf{\textsc{Thought}:} As of October 19, 2025, ASUU has reported progress in its negotiations with the Federal Government. The union's National Strike Coordinating Committee (NSCC) confirmed that the government's negotiation team, led by Alhaji Yayale Ahmed, has presented documented responses and offers regarding key issues. Progress has been noted in: release of third-party deductions; payment of promotion arrears; mainstreaming of Earned Academic Allowances (EAA); resolution of issues concerning the confiscation of University of Abuja land; addressing the victimization of members at KSU, LASU, and FUTO.

The document does not explicitly state what will happen immediately after the current warning strike ends, nor does it confirm that an indefinite strike will follow. Engagements are continuing on the draft renegotiated 2009 Agreement. The union urged its members to remain ``united and resolute in the few days ahead.''

\tcbline

\textbf{Turn 5} \\[2pt]
\textbf{\textsc{Thought}:} As of October 19, 2025, ASUU is in the seventh day of a two-week warning strike. Negotiations are ongoing with the Federal Government's Renegotiation Team, led by Alhaji Yayale Ahmed. ASUU reports that ``progress is being recorded'' on five of its seven-point demands, including the release of third-party deductions, payment of promotion arrears, and the mainstreaming of Earned Academic Allowances (EAA).

The strike is currently in its seventh day. Since the strike began on or around October 13, 2025, it has already met the 168-hour threshold required by the resolution criteria. Because the strike was initiated within the eligible window (October 15--December 31, 2025) and has already reached the 7-day duration requirement, the condition for a ``YES'' resolution has been met.

\tcbline

\textbf{Turn 6} \\[2pt]
\textbf{\textsc{Thought}:} ASUU directed all branches across the country to commence a full withdrawal of services beginning at midnight on Monday, October 13, 2025. The document describes the current action as a ``two-week warning strike.'' While the article notes that the union has previously discussed an ``indefinite strike'' as a potential escalation, it does not explicitly confirm that an indefinite strike will automatically follow the two-week warning period. The union's decision-making process relies on National Executive Council (NEC) meetings and the government's response to their demands.

\tcbline

\textbf{Turn 7} \\[2pt]
\textbf{\textsc{Thought}:} The critical detail here is that the two-week warning strike started on \textbf{October 13, 2025}, which is \textbf{before} the resolution window of October 15, 2025. The criteria explicitly excludes strikes that start before October 15 even if they continue into the window. So the current warning strike doesn't qualify.

The question becomes: Will ASUU declare a \textbf{new} strike (e.g., an indefinite strike) initiated on or after October 15? \\[4pt]

{\color{blue!70}\textbf{\textsc{Search}}:} \texttt{ASUU indefinite strike after warning strike ends October 27 2025} \\
{\color{blue!70}\textbf{\textsc{Search}}:} \texttt{ASUU warning strike ends October 26 what next indefinite} \\
{\color{blue!70}\textbf{\textsc{Search}}:} \texttt{ASUU NEC meeting after warning strike escalation indefinite 2025} \\[2pt]
{\color{green!50!black}\textbf{\textsc{Read}}:} channelstv.com --- \textit{NLC Asks FG To Meet ASUU's Demands Or Risk Strike} \\
{\color{green!50!black}\textbf{\textsc{Read}}:} punchng.com --- \textit{ASUU rejects FG's proposal, begins two-week strike today} \\
{\color{green!50!black}\textbf{\textsc{Read}}:} thenationonlineng.net --- \textit{JUST IN: ASUU declares two-week warning strike}

\tcbline

\textbf{Turn 8} \\[2pt]
\textbf{\textsc{Thought}:} ASUU commenced a two-week warning strike effective from midnight on Monday, October 13, 2025, following the expiration of a 14-day ultimatum issued on September 28, 2025. The document does not explicitly state that an indefinite strike is guaranteed to follow. However, the union's demands---which include the conclusion of the renegotiated 2009 FGN-ASUU agreement, release of withheld salaries, and sustainable funding---remain unresolved. The union leadership rejected the government's recent proposal, describing it as a ``total departure'' from previous agreements. The history of ASUU strikes often involves a progression from warning strikes to indefinite industrial action when negotiations fail.

\tcbline

\textbf{Turn 9} \\[2pt]
\textbf{\textsc{Thought}:} ASUU has declared a two-week warning strike, which commenced on Monday, October 13, 2025. Regarding the escalation sequence, the ASUU National President, Prof.\ Chris Piwuna, stated that the union will meet again after the two-week period. If the government fails to address their demands during this window, the union plans to take ``action that will certainly be different from this one.'' Piwuna explicitly noted that if they embark on the next phase of action, ``it will be total and there will be no going back,'' implying an escalation to an indefinite strike.

\tcbline

\textbf{Turn 10} \\[2pt]
\textbf{\textsc{Thought}:} The Nigeria Labour Congress (NLC) has issued a four-week ultimatum to the Federal Government to meet ASUU's demands. NLC President Joe Ajaero stated that if the government fails to address these issues within this timeframe, the NLC will initiate an indefinite strike in solidarity with educational unions. ASUU has already declared a two-week ``total and comprehensive'' warning strike, which followed the expiration of a 14-day ultimatum on September 28. The document does not explicitly state what happens immediately after this specific two-week warning strike ends.

\tcbline

\textbf{Turn 11} \\[2pt]
{\color{green!50!black}\textbf{\textsc{Read}}:} dailypost.ng --- \textit{Strike: ASUU to hold strategic NEC meeting as action enters week two} \\[2pt]
{\color{blue!70}\textbf{\textsc{Search}}:} \texttt{ASUU NEC meeting October 21 2025 indefinite strike decision} \\
{\color{blue!70}\textbf{\textsc{Search}}:} \texttt{ASUU warning strike ends October 27 next steps indefinite action}

\tcbline

\textbf{Turn 12} \\[2pt]
\textbf{\textsc{Thought}:} The NEC meeting is scheduled to evaluate the union's position on the ongoing strike. ASUU branches held congresses on Monday, October 20, to vote on whether to suspend or continue the current labour action. While the official decision rests with the NEC, reports indicate that many branches have voted in support of continuing the strike. Officials from various universities (University of Abuja, Nnamdi Azikiwe University, and the University of Nigeria, Nsukka) indicated that the matter is ``far from being over'' and that the congresses were largely a formality to confirm support for the strike's continuation. The union is not ``ready to throw in the towel yet.'' The NLC has issued a separate four-week ultimatum to the government.

\tcbline

\textbf{Turn 13} \\[2pt]
{\color{red!70!black}\textbf{\textsc{Final Answer}} (72\%):} The key issue is that ASUU declared a two-week warning strike starting October 13, 2025, which is BEFORE the resolution window begins on October 15. The resolution criteria explicitly state: ``A strike that starts before 2025-10-15 but continues into the window does not qualify.'' Therefore, the current warning strike (Oct 13--27) does NOT count toward resolution.

For YES resolution, ASUU must declare a NEW nationwide strike episode that begins on or after October 15, 2025, and lasts at least 168 consecutive hours (7 days).

The most likely qualifying scenario is that after the two-week warning strike ends ($\approx$October 27), ASUU's NEC declares an indefinite strike as the next escalation step. Key evidence supporting this:

\begin{enumerate}[nosep]
\item ASUU's planned escalation sequence was explicitly: 14-day ultimatum $\to$ 2-week warning strike $\to$ indefinite strike. ASUU National President Prof.\ Chris Piwuna stated the next phase of action ``will be total and there will be no going back.''
\item As of October 21, ASUU is holding a strategic NEC meeting to evaluate the strike, with many branches voting to continue/escalate. Officials from several universities indicated the matter is ``far from being over.''
\item The NLC has issued a 4-week ultimatum threatening solidarity action if ASUU's demands aren't met.
\item The FG has implemented a ``no work, no pay'' policy, which could harden both sides' positions.
\end{enumerate}

Evidence that could prevent escalation:
\begin{enumerate}[nosep]
\item ASUU reports progress in negotiations with FG's Renegotiation Team on 5 of 7 demands.
\item The Senate has intervened to mediate.
\item Historical precedent shows some warning strikes end with partial concessions.
\end{enumerate}

However, ASUU's demands are structural and longstanding (dating to 2009), making full resolution within weeks unlikely. The resolution window extends through December 31, 2025, a very long period that provides multiple opportunities for escalation. Even if the current round of negotiations yields temporary calm, the underlying issues could trigger a new strike at any point during this $\approx$10-week window.

ASUU's historical track record strongly favors escalation: in most recent cycles, warning strikes have been followed by indefinite strikes. The 2022 strike lasted approximately 8 months after similar escalation patterns.

Balancing the strong signals of escalation against the reported negotiation progress and government/Senate intervention, I estimate a probability of approximately 72\%.

\end{tcolorbox}

\subsection{Brazilian Circular Economy Bill}
\label{app:cop30}

\subsubsection*{Forecasting Question}

\begin{tcolorbox}[colback=gray!5, colframe=gray!50]
\small
Will the C\^{a}mara dos Deputados Plenary approve the lead proposition containing PL\,1.874/2022 (National Circular Economy Policy) between 2025-10-15 and 2025-12-31?
\end{tcolorbox}

\subsubsection*{Resolution Criteria}

\begin{tcolorbox}[colback=gray!5, colframe=gray!50, breakable]
\footnotesize
Binary resolution (Yes/No).

\medskip
\textbf{Observation window:} 2025-10-15 00:00:00 UTC through 2025-12-31 23:59:59 BRT. Bras\'{i}lia Time (BRT) is UTC$-$03:00; Brazil has no DST in 2025.

\medskip
\textbf{What counts as YES:} During the observation window, the official C\^{a}mara dos Deputados \textit{ficha de tramita\c{c}\~{a}o} shows that the Plen\'{a}rio approved the proposition that is the \textit{proposi\c{c}\~{a}o principal} (lead) for the package that includes PL\,1.874/2022 at the time of the vote (currently PL\,3.899/2012), with an event explicitly indicating ``Aprovado'' for the proposition in Plenary. If PL\,1.874/2022 is \textit{desmembrada} (de-appended) and itself becomes the item voted in Plenary, a Plenary approval of PL\,1.874/2022 within the window also counts as YES.

\medskip
\textbf{What does NOT count:} Approval of urgency (e.g., REQ~1495/2024) or scheduling alone; approval of a committee opinion without a Plenary approval of the proposition; approval of the \textit{texto-base} alone without subsequent Plenary approval of the proposition.

\medskip
\textbf{How to resolve:}
\begin{enumerate}[nosep,leftmargin=1.5em]
\item Identify the then-current \textit{proposi\c{c}\~{a}o principal} for the package including PL\,1.874/2022 by checking its \textit{ficha de tramita\c{c}\~{a}o}.
\item Open the \textit{ficha} of that lead proposition (as of 2025-10-01, PL\,3.899/2012) and check the 2025 events for a Plenary voting entry indicating ``Aprovado'' during the observation window.
\item If PL\,1.874/2022 itself is voted as principal in the Plenary during the window, check its \textit{ficha} for the Plenary approval event.
\item If the \textit{ficha(s)} explicitly show such a Plenary approval event within the observation window, resolve YES. Otherwise, resolve NO.
\end{enumerate}

\medskip
\textbf{Key context:} PL\,1.874/2022 was presented in the C\^{a}mara on 2024-03-21 and is currently \textit{apensada} under PL\,3.899/2012. The lead proposition's \textit{ficha} shows a relator designation (Dep.\ Luciano Vieira, 2025-03-20), urgency regime metadata, and multiple 2025 plenary entries/schedules. As of late September 2025, there is no recorded Plenary approval.
\end{tcolorbox}

\subsubsection*{Agent Research Trace}

\begin{tcolorbox}[colback=white, colframe=gray!40, breakable]
\footnotesize

\textbf{Turn 1} \\[2pt]
\textbf{\textsc{Thought}:}
\begin{itemize}[nosep,leftmargin=1.5em]
\item Verify the \textit{proposi\c{c}\~{a}o principal} (lead proposition) currently associated with PL\,1.874/2022 on the official C\^{a}mara dos Deputados \textit{ficha de tramita\c{c}\~{a}o}.
\item Ensure the forecast specifically targets the final Plenary approval (``Aprovado'') rather than the approval of urgency (REQ 1495/2024), scheduling, or the \textit{texto-base}.
\item Confirm that the predicted approval event falls within the specific window of October 15, 2025, to December 31, 2025.
\item The final response must contain both a reasoning rationale and a numerical probability score between 0 and 100.
\end{itemize}

\tcbline

\textbf{Turn 2} \\[2pt]
{\color{green!50!black}\textbf{\textsc{Read}}:} camara.leg.br (PL\,1.874/2022 \textit{ficha}) --- \textit{Portal da C\^{a}mara dos Deputados} \\
{\color{green!50!black}\textbf{\textsc{Read}}:} camara.leg.br (PL\,3.899/2012 \textit{ficha}) --- \textit{Portal da C\^{a}mara dos Deputados} \\[2pt]
{\color{blue!70}\textbf{\textsc{Search}}:} \texttt{PL 1874/2022 `economia circular' C\^{a}mara plen\'{a}rio aprovado 2025} \\
{\color{blue!70}\textbf{\textsc{Search}}:} \texttt{PL 3899/2012 plen\'{a}rio aprovado 2025}

\tcbline

\textbf{Turn 3} \\[2pt]
\textbf{\textsc{Thought}:} The ``Situa\c{c}\~{a}o'' field lists PL\,1.874/2022 as \textit{apensado} to PL\,1755/2022. However, the most recent legislative action (20/03/2025) identifies the bill as \textit{apensado} to PL\,3.899/2012. Given that the latter is the most recent update, PL\,3.899/2012 is the current lead proposition.

Status: ``Aguardando Cria\c{c}\~{a}o de Comiss\~{a}o Tempor\'{a}ria pela MESA; Pronta para Pauta no Plen\'{a}rio (PLEN).''

The bill has not been approved in the Plenary. Active \textit{desapensa\c{c}\~{a}o} requests (REQ\,3873/2025, REQ\,4116/2025) are pending. No scheduled Plenary vote recorded.

\tcbline

\textbf{Turn 4} \\[2pt]
\textbf{\textsc{Thought}:} PL\,3.899/2012 remains active and is currently subject to Plenary appreciation. It has \textbf{not} been approved in the Plenary in 2025.

Most recent events: 2025-10-15 PRLP~n.~8 PLEN by Dep.\ Luciano Vieira; 2025-10-08 PRLP~n.~7 PLEN and Plenary session where matter not appreciated; 2025-10-02 REQ\,4116/2025 (\textit{desapensa\c{c}\~{a}o}); 2025-09-26 PRLP~n.~6 PLEN; 2025-09-24 REQ\,3873/2025 (\textit{desapensa\c{c}\~{a}o}).

Throughout 2025, the proposition has been scheduled for multiple Plenary sessions but in every instance: ``Mat\'{e}ria n\~{a}o apreciada em face do encerramento da Sess\~{a}o'' or ``Mat\'{e}ria n\~{a}o apreciada em face do encerramento da Ordem do Dia.''

\tcbline

\textbf{Turn 5} \\[2pt]
{\color{blue!70}\textbf{\textsc{Search}}:} \texttt{PL 3899/2012 economia circular plen\'{a}rio vota\c{c}\~{a}o outubro 2025} \\
{\color{blue!70}\textbf{\textsc{Search}}:} \texttt{economia circular c\^{a}mara plen\'{a}rio vota\c{c}\~{a}o 2025 aprovado} \\[2pt]
{\color{green!50!black}\textbf{\textsc{Read}}:} politicabrasileira.com.br --- \textit{Projeto sobre economia circular deve ser pauta de reuni\~{a}o de l\'{i}deres da C\^{a}mara} \\
{\color{green!50!black}\textbf{\textsc{Read}}:} blogs.correiobraziliense.com.br --- \textit{PL da Economia Circular virou `monstro legislativo', dizem representantes do setor produtivo}

\tcbline

\textbf{Turn 6} \\[2pt]
\textbf{\textsc{Thought}:} As of September 11, 2025, PL\,3899/2012 is facing significant opposition from the productive sector, which characterizes the current substitute text as a ``legislative monster.'' Obstacles: the bill has been distorted by the attachment (\textit{apensa\c{c}\~{a}o}) of dozens of other projects, resulting in a text with over 100 articles; industry argues the text imposes unrealistic obligations (mandatory Circularity Plans, changes to energy matrices) and disproportionate sanctions (fines up to 4\% of gross revenue, capped at R\$\,100 million, product seizures, license suspensions); the sector actively supports PL\,1874/2022 instead of the current substitute for PL\,3899/2012.

\tcbline

\textbf{Turn 7} \\[2pt]
\textbf{\textsc{Thought}:} As of September 13, 2025, the project was expected to be brought to a meeting of party leaders by the President of the Chamber. The Chamber had previously scheduled the project for a Plenary vote in the week prior to September 13, 2025, but parliamentarians did not have sufficient time to deliberate on it. There is no specific date provided for a future vote, only that it is considered a priority. Finance Minister Fernando Haddad cited it as a government priority.

\tcbline

\textbf{Turn 8} \\[2pt]
{\color{blue!70}\textbf{\textsc{Search}}:} \texttt{PL 3899 economia circular plen\'{a}rio outubro 2025 vota\c{c}\~{a}o} \\
{\color{blue!70}\textbf{\textsc{Search}}:} \texttt{economia circular c\^{a}mara deputados vota\c{c}\~{a}o plen\'{a}rio outubro novembro 2025} \\
{\color{blue!70}\textbf{\textsc{Search}}:} \texttt{c\^{a}mara deputados pauta plen\'{a}rio semana outubro 2025}

\tcbline

\textbf{Turn 9} \\[2pt]
{\color{blue!70}\textbf{\textsc{Search}}:} \texttt{PL 3899 economia circular vota\c{c}\~{a}o plen\'{a}rio c\^{a}mara outubro novembro 2025} \\
{\color{blue!70}\textbf{\textsc{Search}}:} \texttt{economia circular c\^{a}mara aprovado plen\'{a}rio site:camara.leg.br 2025} \\[2pt]
{\color{green!50!black}\textbf{\textsc{Read}}:} camara.leg.br (PL\,3.899/2012 \textit{ficha}, re-read) --- \textit{Portal da C\^{a}mara dos Deputados}

\tcbline

\textbf{Turn 10} \\[2pt]
\textbf{\textsc{Thought}:} October 2025 events on the PL\,3.899/2012 \textit{ficha}: 02/10/2025 REQ\,4116/2025 (\textit{desapensa\c{c}\~{a}o}) by Dep.\ Elcione Barbalho; 07/10/2025 ``Mat\'{e}ria n\~{a}o apreciada em face do encerramento da Ordem do Dia''; 08/10/2025 same, plus PRLP~n.~7 by Dep.\ Luciano Vieira; 15/10/2025 PRLP~n.~8 by Dep.\ Luciano Vieira.

The matter has \textbf{not} been voted or approved in Plenary in October 2025.

\tcbline

\textbf{Turn 11} \\[2pt]
{\color{blue!70}\textbf{\textsc{Search}}:} \texttt{PL 3899 economia circular aprovado c\^{a}mara outubro 2025} \\
{\color{blue!70}\textbf{\textsc{Search}}:} \texttt{c\^{a}mara deputados pauta plen\'{a}rio economia circular semana}

\tcbline

\textbf{Turn 12} \\[2pt]
{\color{green!50!black}\textbf{\textsc{Read}}:} ndmais.com.br --- \textit{Sustentabilidade: governo, oposi\c{c}\~{a}o e setor produtivo fecham acordo `hist\'{o}rico'} \\[2pt]
{\color{blue!70}\textbf{\textsc{Search}}:} \texttt{economia circular acordo c\^{a}mara vota\c{c}\~{a}o plen\'{a}rio site:camara.leg.br OR site:agenciacamara.leg.br 2025} \\
{\color{blue!70}\textbf{\textsc{Search}}:} \texttt{economia circular acordo governo oposi\c{c}\~{a}o c\^{a}mara vota\c{c}\~{a}o outubro 2025}

\tcbline

\textbf{Turn 13} \\[2pt]
\textbf{\textsc{Thought}:} On Wednesday, September 24, 2025, a ``historic'' consensus was reached between the government, the opposition, and various sectors (including agribusiness, industry, and environmentalists) to de-append (\textit{desapensar}) PL\,1.874/2022 from PL\,3.899/2012. The agreement means PL\,1.874/2022---which has already been approved by the Senate---will be separated from the older PL\,3.899/2012, allowing it to proceed as the primary text for a vote in the Plenary. The decision to de-append rests with the President of the Chamber, Hugo Motta. The expectation is for the matter to be voted under an urgency regime. No specific date for the Plenary vote was provided.

\tcbline

\textbf{Turn 14} \\[2pt]
{\color{blue!70}\textbf{\textsc{Search}}:} \texttt{PL 1874 OR PL 3899 economia circular desapensa\c{c}\~{a}o c\^{a}mara outubro 2025} \\
{\color{blue!70}\textbf{\textsc{Search}}:} \texttt{economia circular c\^{a}mara deputados plen\'{a}rio vota\c{c}\~{a}o semana outubro 2025} \\[2pt]
{\color{green!50!black}\textbf{\textsc{Read}}:} camara.leg.br (PL\,1.874/2022 \textit{ficha}, re-read) --- \textit{Portal da C\^{a}mara dos Deputados}

\tcbline

\textbf{Turn 15} \\[2pt]
\textbf{\textsc{Thought}:} October 2025: one entry recorded---02/10/2025 REQ\,4116/2025 by Dep.\ Elcione Barbalho requesting \textit{desapensa\c{c}\~{a}o}. The bill has \textbf{not} been de-appended. Status remains ``Apensado ao PL\,1755/2022.'' Two requests pending (REQ\,3873/2025 on 24/09/2025 and REQ\,4116/2025 on 02/10/2025). No record of Plenary vote. The bill is marked ``Pronta para Pauta no Plen\'{a}rio'' but the ongoing procedural debate regarding its attachment status makes a Plenary approval uncertain.

\tcbline

\textbf{Turn 16} \\[2pt]
{\color{blue!70}\textbf{\textsc{Search}}:} \texttt{economia circular OR PL 3899 OR PL 1874 c\^{a}mara vota\c{c}\~{a}o plen\'{a}rio 2025 after:2025-10-15} \\
{\color{blue!70}\textbf{\textsc{Search}}:} \texttt{c\^{a}mara deputados plen\'{a}rio pauta vota\c{c}\~{a}o semana 20 outubro 2025}

\tcbline

\textbf{Turn 17} \\[2pt]
{\color{red!70!black}\textbf{\textsc{Final Answer}} (30\%):} PL\,1.874/2022 (National Circular Economy Policy) is currently \textit{apensada} to PL\,3.899/2012, which serves as the lead proposition. As of October 15, 2025, PL\,3.899/2012 has NOT been approved in Plenary. Throughout 2025, the bill has been repeatedly scheduled for Plenary sessions but consistently not voted, with entries showing ``Mat\'{e}ria n\~{a}o apreciada em face do encerramento da Ordem do Dia.'' The rapporteur (Dep.\ Luciano Vieira) has been actively filing preliminary opinions, up to PRLP~8 as of October 15, 2025.

Several factors support potential approval:
\begin{enumerate}[nosep]
\item The bill has urgency status since November 2024.
\item Finance Minister Haddad cited it as a government priority.
\item A ``historic'' agreement was reached on September 24, 2025, between government, opposition, and productive sectors to de-append PL\,1.874/2022 from PL\,3.899/2012, potentially simplifying its path to a vote.
\item End-of-year legislative sessions typically see more bills cleared.
\end{enumerate}

However, significant obstacles remain:
\begin{enumerate}[nosep]
\item The de-appending has NOT yet been effectuated---requests REQ~3873/2025 and REQ~4116/2025 are still pending.
\item The productive sector called the current substitute text a ``legislative monster'' with over 100 articles and unrealistic obligations.
\item The bill has failed to reach a vote in every Plenary session it was scheduled for in 2025.
\item The procedural path forward remains uncertain due to pending de-appending requests.
\end{enumerate}

While there is political will and approximately 2.5 months remain in the observation window, the repeated failure to vote despite being on the agenda, ongoing procedural uncertainty about de-appending, and substantial industry opposition make approval uncertain but possible given end-of-year legislative dynamics.

\end{tcolorbox}

\end{document}